\documentclass[conference]{IEEEtran}
\IEEEoverridecommandlockouts
\usepackage{cite}
\usepackage{amsmath,amssymb,amsfonts}
\usepackage{algorithmic}
\usepackage{graphicx}
\usepackage{textcomp}
\usepackage{xcolor}
\usepackage{subcaption}
\usepackage{array, tabularray, multirow}
\usepackage{hyperref}
\usepackage{cleveref}

\DeclareMathOperator{\diag}{diag} 

\def\BibTeX{{\rm B\kern-.05em{\sc i\kern-.025em b}\kern-.08em
    T\kern-.1667em\lower.7ex\hbox{E}\kern-.125emX}}
\begin{document}

\title{\LARGE \bf Shape Manipulation of Bevel-Tip Needles for Prostate Biopsy Procedures: A Comparison of Two Resolved-Rate Controllers}
\author{Yanzhou Wang$^{^\dagger{1}}$, Lidia Al-Zogbi$^{^\dagger{1}}$, Jiawei Liu$^{2}$, Lauren Shepard$^{3}$, Ahmed Ghazi$^{3}$, Junichi Tokuda$^{4}$, \\Simon Leonard$^{5}$, Axel Krieger$^{^\ddagger{1}}$, and Iulian Iordachita$^{^\ddagger{1}}$ 
\thanks{*This work is supported by NIH R01EB020667, 1R01EB025179, 1R01CA235134, and in part by a collaborative research agreement with the Multi-Scale Medical Robotics Center in Hong Kong.}
\thanks{$^\dagger$Joint first authors. $^\ddagger$Joint last authors.}
\thanks{$^{1}$Yanzhou Wang, Lidia Al-Zogbi, Axel Krieger, and Iulian Iordachita are with the Department of Mechanical Engineering and the Laboratory for Computational Sensing and Robotics, Johns Hopkins University, Baltimore, MD, USA
  {\tt\small ywang521@jh.edu}}%
\thanks{$^{2}$Jiawei Liu is with the Laboratory of Computational Sensing and Robotics, Johns Hopkins University, Baltimore, MD, USA}
\thanks{$^{3}$Lauren Shepard and Ahmed Ghazi are with the Brady Urological Institute, Johns Hopkins University, Baltimore, MD, USA}
\thanks{$^{4}$Junichi Tokuda is with the Department of Radiology, Brigham and Women’s Hospital and Harvard Medical School, Boston, MA, USA}
\thanks{$^{5}$Simon Leonard is with the Department of Computer Science and the Laboratory for Computational Sensing and Robotics, Johns Hopkins University, Baltimore, MD, USA}
}

\onecolumn
\noindent This work has been submitted to the IEEE for possible publication. Copyright may be transferred without notice, after which this version may no longer be accessible.

\twocolumn

\maketitle

\begin{abstract}
Prostate cancer diagnosis continues to encounter challenges, often due to imprecise needle placement in standard biopsies. Several control strategies have been developed to compensate for needle tip prediction inaccuracies, however none were compared against each other, and it is unclear whether any of them can be safely and universally applied in clinical settings. This paper compares the performance of two resolved-rate controllers, derived from a mechanics-based and a data-driven approach, for bevel-tip needle control using needle shape manipulation through a template. We demonstrate for a simulated 12-core biopsy procedure under model parameter uncertainty that the mechanics-based controller can better reach desired targets when only the final goal configuration is presented even with uncertainty on model parameters estimation, and that providing a feasible needle path is crucial in ensuring safe surgical outcomes when either controller is used for needle shape manipulation.
\end{abstract}


\section{Introduction}
\label{sec:introduction}

Prostate cancer remains a significant global health challenge, being the most commonly diagnosed cancer and the second leading cause of cancer-related deaths among men \cite{bashir2015epidemiology}. Despite recent advancements in diagnostic techniques, particularly in Magnetic Resonance Imaging (MRI) and Ultrasound (US)-guided prostate biopsies, the false negative rate for a standard 12-core biopsy still exceeds 30\%, a problem that persists even with repeated interventions \cite{serefoglu2013reliable}. This issue is compounded by inherent undersampling, emphasizing the need for precise placement of the biopsy needle to effectively mitigate diagnostic challenges \cite{presti2007prostate}. The susceptibility to errors in prostate biopsies primarily stems from two factors: the challenges associated with controlling the tip position of the flexible biopsy needle, and the movement of targets within the prostate during the procedure. This manuscript focuses on the first challenge -- flexible needle control -- under the assumption that the location of target points can be tracked.

To study the effect of bevel angle on a needle tip's trajectory, research efforts have focused on three main modeling approaches: kinematic, finite element, and mechanics-based~\cite{babaiasl2022robotic}. Kinematic models assume a vehicle-type system dynamics for the needle tip, with the flexible needle shaft following the tip's path. However, this assumption does not always hold true, especially with higher gauge needles~\cite{webster2006nonholonomic, reed2011robot}. Finite element models comprehensively incorporate needle-tissue interactions, and can achieve high accuracy with proper tuning. However, they face challenges linked to computational intensity, as well as their dependence on a range of physical and solver-specific parameters~\cite{adagolodjo2019robotic, Terzano2020}. Mechanics-based models present a more effective solution by directly coupling beam mechanics with tissue reaction forces, which can be adapted to various levels of simplicity or complexity depending on specific use cases~\cite{lehmann2018model, wang2023flexible}. Mechanics-based methods not only alleviate the computational burden of full material deformation modeling seen in the case of finite element approaches, but also better capture needle bending behavior in response to tissue compression, which is overlooked by kinematic models.

Mechanics-based models in needle shape prediction still face accuracy limitations due to their dependency on mechanical and geometrical properties of the needle and tissues. This issue can be mitigated by introducing closed-loop controls that use feedback to compensate for modeling inaccuracies. For instance, Khadem \textit{et al.}~\cite{khadem2016ultrasound} developed a nonlinear model predictive needle controller integrated with US-based needle tip position feedback, achieving a maximum targeting error of 2.85mm in tissue steering experiments, albeit only by controlling the needle's axial rotation. With the increased usage of transperineal biopsy templates~\cite{abdulmajed2015role}, omitting lateral needle movement limits the clinical translation of such approach. Lehmann \textit{et al.}~\cite{lehmann2017deflection} developed a semi-autonomous needle insertion platform for US-guided prostate brachytherapy, modeling a point force application to an inserted bevel-tip needle for control using beam mechanics. The authors validated their model using US tracking and force measurements with an average targeting error of roughly 2mm for a 140mm needle insertion depth. However, neither studies \cite{lehmann2017deflection, khadem2016ultrasound} evaluate the methods' effectiveness when model parameters do not precisely match the physical setup, as is often the case in practice. 


\begin{figure*}[t]
  \centering
  \includegraphics[width=\textwidth]{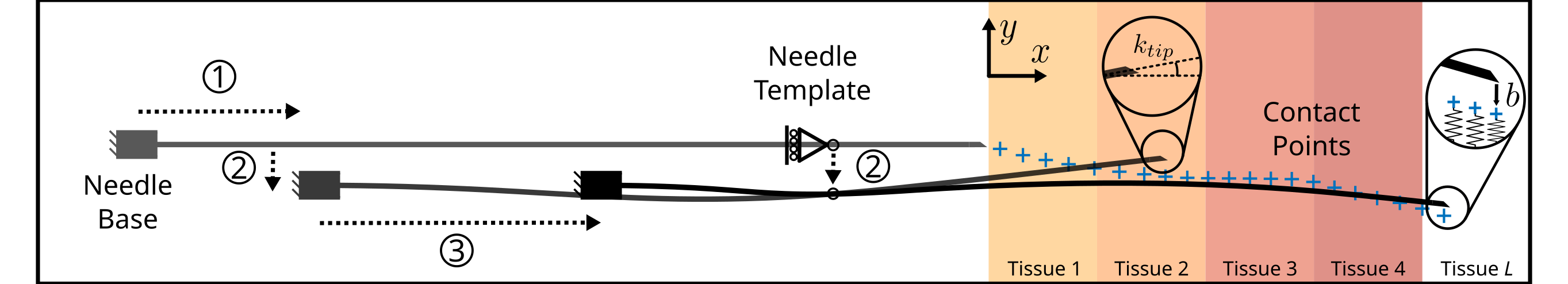}
  \caption{Schematic of bevel-tip needle insertion into multi-layered soft tissues. The global frame is placed at needle entry point on skin surface. Needle base translates along the $x$- and $y$-axis with a fixed slope; the needle template moves independently along the $y$-axis with an unconstrained slope. Bevel effect can be achieved by placing contact points (blue crosses) to drive the needle tip in the direction of the bevel ($-y$ in this case). Final needle shape is achieved through a sequence of base and template motions \textcircled{1} through \textcircled{3}.}
  \label{fig:schematic}
\end{figure*}

Data-driven strategies can present as an alternative to modeling~\cite{bernardes2023data, simon_icra2024}. For example, utilizing Broyden's method, Bernardes \textit{et al.}~\cite{bernardes2023data} used a bevel-tip needle with an embedded electromagnetic tracking sensor for collaborative, robot-assisted prostate biopsies with a needle-guiding stage. By updating the mapping between actuator inputs and observed changes in needle tip position, the authors were able to achieve targeting accuracy of less than 1mm in \textit{ex vivo} tissue. Ma \textit{et al.}~\cite{simon_icra2024} extended Bernardes' work to develop a data-driven controller for fully autonomous bevel-tip needle steering. However, despite the promising results achieved by data-driven strategies, it is unclear whether they can be safely and universally applied in clinical settings, and how their performance compares to a mechanics-based strategy when such model exists.


In this manuscript, we formulate the mechanics of local needle shape changes along the needle shaft when interacting with nonlinear soft tissues, and examine the effect of shape manipulation on a bevel-tip needle's trajectory within a simulated environment. Specifically, we compare two resolved-rate controllers whose local system dynamics are derived numerically via either a data-driven or a mechanics-based method. We demonstrate that the mechanics-based controller can better reach desired targets when only the final goal configuration is presented even with uncertainty on model parameters estimation, and that providing a feasible needle path is crucial in ensuring safe surgical outcomes when either controller is used for needle shape manipulation.

The rest of the paper is structured as follows: Sec.~\ref{sec:materials_and_methods} introduces a simulator for comparing controller performances, a prostate phantom for collecting simulator data, and two formulations for approximating local system dynamics. Sec.~\ref{sec:results_and_discussion} presents the comparison results when performing path-following and point-stabilization tasks, and discusses shape manipulation as a general strategy for needle control. Sec.~\ref{sec:conclusion_and_future_work} summarizes our findings.

\section{Materials and Methods}
\label{sec:materials_and_methods}

\subsection{Bevel-Tip Needle Shape Manipulation and Simulation}
\label{sec:mechanics_based_model}
An interactive simulator developed by authors in~\cite{wang2023flexible, wang_icra2024} is used to compare the performance of mechanics-based and data-driven controllers. In the planar case shown in Fig.~\ref{fig:schematic}, the simulator treats the needle as an inextensible Euler-Bernoulli beam, and the multi-layered soft tissue environment as incompressible, Ogden-type hyperelastic materials. Under shape manipulation, the needle experiences bending forces due to displacement inputs as well as interaction with soft tissues.

Tissue reaction force is nonlinear and exhibits strain-hardening behavior under compression. As proposed in~\cite{wang2023flexible}, in a multi-layered insertion scenario, the total force $F$ is calculated by aggregating the contributed forces from individual layers through
\begin{align}
  \label{eq:bending_force}
  &F(x)  = \sum_i^Lf_i(x) \quad x\in\Omega, \\ 
 \label{eq:individual_force}
f_i(x) &\approx k_i(x)u(x)\quad x \in \Omega_i \subseteq \Omega,
\end{align}
where $u(x)$ is the deformed shape of the needle, and subscript $i$ relates to quantities pertaining to the $i^{\text{th}}$ tissue layer. Tissue tangent stiffness $k$ increases with degree of compression through
\begin{equation}
  \label{eq:tangent_modulus}
  k(\lambda) = \frac{\partial\sigma_{comp}}{\partial\lambda} =  2\mu \left( \lambda^{\alpha-1} + \frac{1}{2}\lambda^{-\frac{\alpha}{2}-1} \right),
\end{equation}
where $\mu$ is the shear modulus, $\alpha$ a nonlinearity measure, and $\sigma_{comp}$ the stress component in the direction of tissue compression. The stretch variable $\lambda$ is defined as
\begin{equation}
  \label{eq:stretch_ratio}
  \lambda = \frac{t_i - |u(x)|}{t_i},
\end{equation}
where $t_i$ is the tissue's initial thickness prior to compression. \Cref{eq:individual_force,eq:tangent_modulus,eq:stretch_ratio} describe the strain-hardening behavior as the tissue becomes more compressed by the needle.

To simulate the needle's bevel effect during insertion, offset contact points are added iteratively at the needle tip based on the depth of insertion. These offset contact points act as ``pre-compressed springs'' such that the needle naturally bends towards the direction of the bevel. The value of the offset, $b$ (see Fig.~\ref{fig:schematic}), can be modified to create a bevel effect of varying direction and magnitude~\cite{wang_icra2024}. Static equilibrium of Euler-Bernoulli beam bending equation subject to distributed tissue forces is solved using a nonlinear finite element routine, which considers the effect of contact points on a per-element basis. Needle displacement inputs are treated as essential boundary conditions in the simulator. See \cite{wang_icra2024} for more details.

In the insertion example shown in Fig.~\ref{fig:schematic}, the needle first passes through a template before reaching soft tissues. Displacement inputs to the needle are placed at the needle base and needle template. The needle base is modeled as a fixed support at each time step and is allowed to translate along the $x$- and $y$-axis, while the needle template acts as a roller support and only moves along the $y$-axis. Such physical constraints are enforced as essential boundary conditions in the finite element solver, and can be modified depending on the specific hardware setup. By coupling needle base and template motions with bevel-tip insertion behavior, needle shape with multi-step inputs can be simulated.

\subsection{Parameter Tuning with Prostate Phantom}
\label{sec:parameter_tuning}
\begin{figure}[t]
\begin{subfigure}{.68\columnwidth}
  \centering
  \includegraphics[width=1\columnwidth]{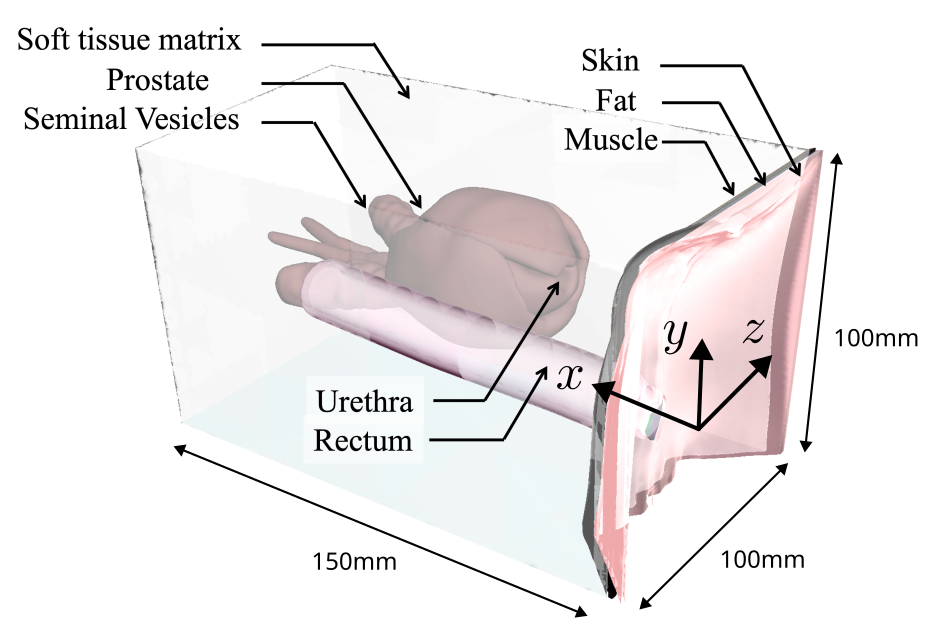}  
  \caption{}
  \label{fig:sub-first}
\end{subfigure}
\begin{subfigure}{.28\columnwidth}
  \centering
  \includegraphics[width=1\columnwidth]{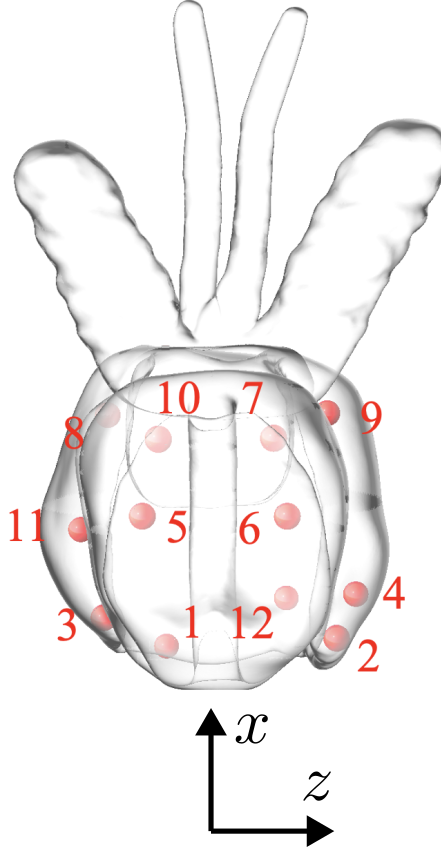}  
  \caption{}
  \label{fig:sub-second} 
\end{subfigure}
\caption{(a) Prostate phantom with internal organs. (b) Selected 12-core biopsy target points in prostate phantom. Direction $z$ is flattened for current planar insertion scenario.}
\label{fig:phantom}
\end{figure}

Since the comparative study between two controllers is carried out in simulation, it is important that the identified simulator parameters be representative of a real physical object. For our study, we utilize a patient-specific prostate phantom~\cite{stone2023systems, saba2022design}. As shown in Fig.~\ref{fig:phantom}, the phantom consists of a prostate, urethra, seminal vesicles, and rectum, submerged in a volume of surrounding soft tissue. Additional layers of muscle, fat, and skin are added to the rectal side of the phantom. All organs are created using 3D printed negative molds filled with hydrogels. Different organs have varied hydrogel contents composition to match tensile strength properties obtained from cadaveric samples \cite{saba2023three}. The molds are obtained from expert-segmented anonymized MRI scans of a male patient. A total of 6 clinical experts have evaluated the phantom, and testified to its realism as a training tool for prostate biopsies \cite{shepard2024hydrogel}. 

Tissue parameters $\alpha$ and $\mu$ (see \Cref{eq:tangent_modulus}), as well as contact point offset $b$ (see Fig. \ref{fig:schematic}), are obtained through iterative insertion simulation until the final resulting needle shape matches data collected on the physical phantom. Tuned parameters as well as average layer thickness, denoted by $\Delta\Omega$, for each phantom tissue layer are reported in Table~\ref{tab:params} and are used in the simulator to represent the prostate phantom. The contact point offset $b$ was identified to be $-0.3$. Note that these values are representative of the phenomenological behavior of the needle in simulation rather than actual physical properties of tissue materials. Different sets of values can recreate similar needle behavior, but this aspect is not the focus of this manuscript.

\begin{table}[b]
\centering
\caption{Tuned model parameters and average layer thicknesses.}
\label{tab:params}
\resizebox{\columnwidth}{!}{%
\begin{tabular}{|cc|cc|cc|cc|cc|}
\hline
\multicolumn{2}{|c|}{\textbf{Skin}} & \multicolumn{2}{c|}{\textbf{Fat}} & \multicolumn{2}{c|}{\textbf{Muscle}} & \multicolumn{2}{c|}{\textbf{Soft Tissue}} & \multicolumn{2}{c|}{\textbf{Prostate}} \\ \hline
\multicolumn{1}{|c|}{$\mu$} & $\alpha$ & \multicolumn{1}{c|}{$\mu$} & $\alpha$ & \multicolumn{1}{c|}{$\mu$} & $\alpha$ & \multicolumn{1}{c|}{$\mu$} & $\alpha$ & \multicolumn{1}{c|}{$\mu$} & $\alpha$ \\ \hline
\multicolumn{1}{|c|}{2e2MPa} & 1 & \multicolumn{1}{c|}{1.5e1MPa} & -1 & \multicolumn{1}{c|}{3e1MPa} & -1 & \multicolumn{1}{c|}{8e2MPa} & -1 & \multicolumn{1}{c|}{3.2e3MPa} & 1 \\ \hline
\multicolumn{2}{|c|}{$\Delta\Omega$: 2mm} & \multicolumn{2}{c|}{$\Delta\Omega$: 3mm} & \multicolumn{2}{c|}{$\Delta\Omega$: 5mm} & \multicolumn{2}{c|}{$\Delta\Omega$: 10.5mm} & \multicolumn{2}{c|}{$\Delta\Omega$: 55mm} \\ \hline
\end{tabular}%
}
\end{table}

\subsection{Resolved-Rate Controllers for Needle Control}
\label{sec:resolved_rate_controllers}
A resolved-rate controller relies on a system dynamics that relates input and output velocities with a Jacobian, $\mathbf{J}$:
\begin{equation}
\dot{\mathbf{y}} = \mathbf{J}(\mathbf{x})\dot{\mathbf{x}}.\label{eq:resolved_rate_dynamics}
\end{equation}
For tracking and regulation, the following discretized control law can be used:
\begin{equation}
  \Delta \mathbf{x}^{[n + 1]} = -\left(\mathbf{J}^{[n]}\right)^{-1}\mathbf{K_p}(\mathbf{y}^{[n]} - \mathbf{y_d}^{[n]}), \label{eq:control_law}
\end{equation}
where $\Delta\mathbf{x}$ is the control input, $\mathbf{K_p}$ is a proportional gain matrix, and $\mathbf{y_d}$ is the desired output. The superscript $n$ denotes the $n^\text{th}$ time step. This control law necessitates repetitive updates and the inversion of the Jacobian matrix, under the assumption that $\left(\mathbf{J}^{[n]}\right)^{-1}$ indeed exists.

In the present case, the positions of needle base and template are the three control inputs to the system, \textit{i.e.} $\mathbf{x} = [x_{base}, y_{base}, y_{template}]^{\top}$, and the needle tip position and slope as three system outputs, \textit{i.e.} $\mathbf{y} = [x_{tip}, y_{tip}, k_{tip}]^{\top}$ .

The goal is to obtain a mapping $\mathbf{J}$ between the three input-output sets to implement a resolved-rate control law. Here, we describe two methods of obtaining and updating $\mathbf{J}$ -- a data-driven method and a mechanics-based method.

\subsubsection{Data-driven Method}
\label{sec:data_driven_control}
A data-driven approach relies on an update law to obtain and update the Jacobian matrix. Using Broyden's method~\cite{broyden1965class}, the Jacobian matrix can be iteratively approximated based on the following rule:
\begin{equation}
  \mathbf{J}^{[n]} = \mathbf{J}^{[n - 1]} + \frac{\mathbf{\Delta y}^{[n]} - \mathbf{J}^{[n - 1]}\Delta \mathbf{x}^{[n]}}{\Vert \Delta \mathbf{x}^{[n]}\Vert ^2} \left(\Delta \mathbf{x}^{[n]}\right)^{\top}. \label{eq:update_law}
\end{equation}
where $\Delta \mathbf{y}^{[n]} = \mathbf{y}^{[n]} - \mathbf{y}^{[n - 1]}$ is the difference in function value at the current and previous iterations, and $\Delta \mathbf{x}^{[n]}$ is the difference in function arguments at the current and previous iteration. 

\subsubsection{Mechanics-based Method}
\label{sec:mechanics_based_control}
A mechanics-based approach relies on a physical or phenomenological model and simulation of the system. In the case of a bevel-tip needle insertion through a movable template, by obtaining a static equilibrium solution of the model presented in Sec.~\ref{sec:mechanics_based_model}, the input-output system can be written as
\begin{equation}
\mathbf{y} = \mathbf{f}(\mathbf{x}),
\end{equation}
where $\mathbf{x} \in \mathcal{R}^N$ denotes the input to the solver, such as a prescribed nodal displacement, and $\mathbf{y} \in \mathcal{R}^M$ denotes the output, such as nodal position and orientation.

Instead of an update law such as~\Cref{eq:update_law} that requires feedback data $\mathbf{y}^{[n]}$ from the physical system, the mechanics-based method can obtain the input-output Jacobian by simulating changes in system outputs in response to small changes in inputs at each time step. Using finite difference methods, such as the central difference method, the Jacobian can be obtained numerically:
\begin{equation}
  \mathbf{J}_{ij}^{[n]} = \frac{\mathbf{f}_i^{[n]}(\mathbf{x}_j^{[n]} + \epsilon) - \mathbf{f}_i^{[n]}(\mathbf{x}_j^{[n]} - \epsilon)}{2\epsilon},\label{eq:central_difference}
\end{equation}
where we use the indicial notation to denote element index. Indices $i \in M$ and $j \in N$, and constant $\epsilon$ is a small number. 
The input-output relation in terms of their variation can be written as
\begin{equation}
\delta \mathbf{y} = \mathbf{J}(\mathbf{x})\delta \mathbf{x}.\label{eq:variation}
\end{equation}
Although not strictly necessary, we focus our attention on the present case with $M = N = 3$, \textit{i.e.} when $\mathbf{J}$ is a square matrix. 

\subsection{Simulated Targeting Experiments}
\label{sec:targeting_experiments}
Comparison between data-driven and mechanics-based needle controllers is carried out in the simulator (Sec.\ref{sec:mechanics_based_model}). A total of 12 target points within the prostate phantom are selected (Fig.~\ref{fig:sub-second}), emulating a 12-core biopsy procedure. For each target, we compare the two controllers in both path-following and point-stabilization tasks, \textit{i.e.} retracing a desired needle tip path from entry to target, and reaching a target point without path guidance using only the start and end needle configurations. The viable paths for the needle tip are generated in the simulator through forward needle insertion (step \textcircled{1} in Fig.~\ref{fig:schematic}). Upon reaching the target depth, the deflection of the needle tip is recorded and used as an initial offset at the entry point, such that by simply moving the needle in the $+x$-direction with zero initial slope, the needle tip reaches the selected target without $y$-axis adjustments.

According to~\cite{Kelly2019}, cancerous and non-cancerous prostate tissues exhibit substantial differences in their elastic moduli. 
Most reported standard deviations fall approximately within $30-50\%$ of the mean elastic moduli for benign and cancerous prostate tissues (see Fig. 4 of~\cite{Kelly2019}). In order to simulate scenarios when patient-specific tissue parameters differ from tuned parameter values, for mechanics-based controllers, we consider $\pm50\%$ parametric variation for each tissue layer based on Table~\ref{tab:params}. Identical proportional gain $\mathbf{K_p} = \diag{[0.5, 1, 0.1]}$ are set for both controllers. The simulated needle is an $18$ gauge, $170$mm-long solid bevel-tip needle. Needle template is positioned $22$mm away from the skin.

Although needle tip slope $k_{tip}$ is used as an output, since the slope is less clinically relevant in assessing the surgical outcome, control execution stops when the final needle tip position error $err \leq 0.25$mm from the target, where

\begin{equation}
  \label{eq:error_mag}
  err = \sqrt{(x_{tip} - x_{goal})^2 + (y_{tip} - y_{goal})^2}.
\end{equation}

As a metric for controller performance evaluation, for each target and task, we calculate the percentage of maximum shape manipulation effort over final target depth:
\begin{align}
  \label{eq:metric}
  P_{(\cdot)} = \frac{\max{\vert{\Delta y_{(\cdot)}}\vert}}{x_{target}} \times 100,
\end{align}
where subscript $(\cdot)$ denotes either needle base or needle template. Since for each target, $x_{target}$ is fixed, a smaller $P$ value indicates a smaller amount of adjustment for that specific controller and task.

\section{Results and Discussion}
\label{sec:results_and_discussion}



\subsection{Controller Performance Comparison}
\label{sec:controller_performance_comparison}
\begin{figure}[t]
  \centering
  \includegraphics[width=\columnwidth]{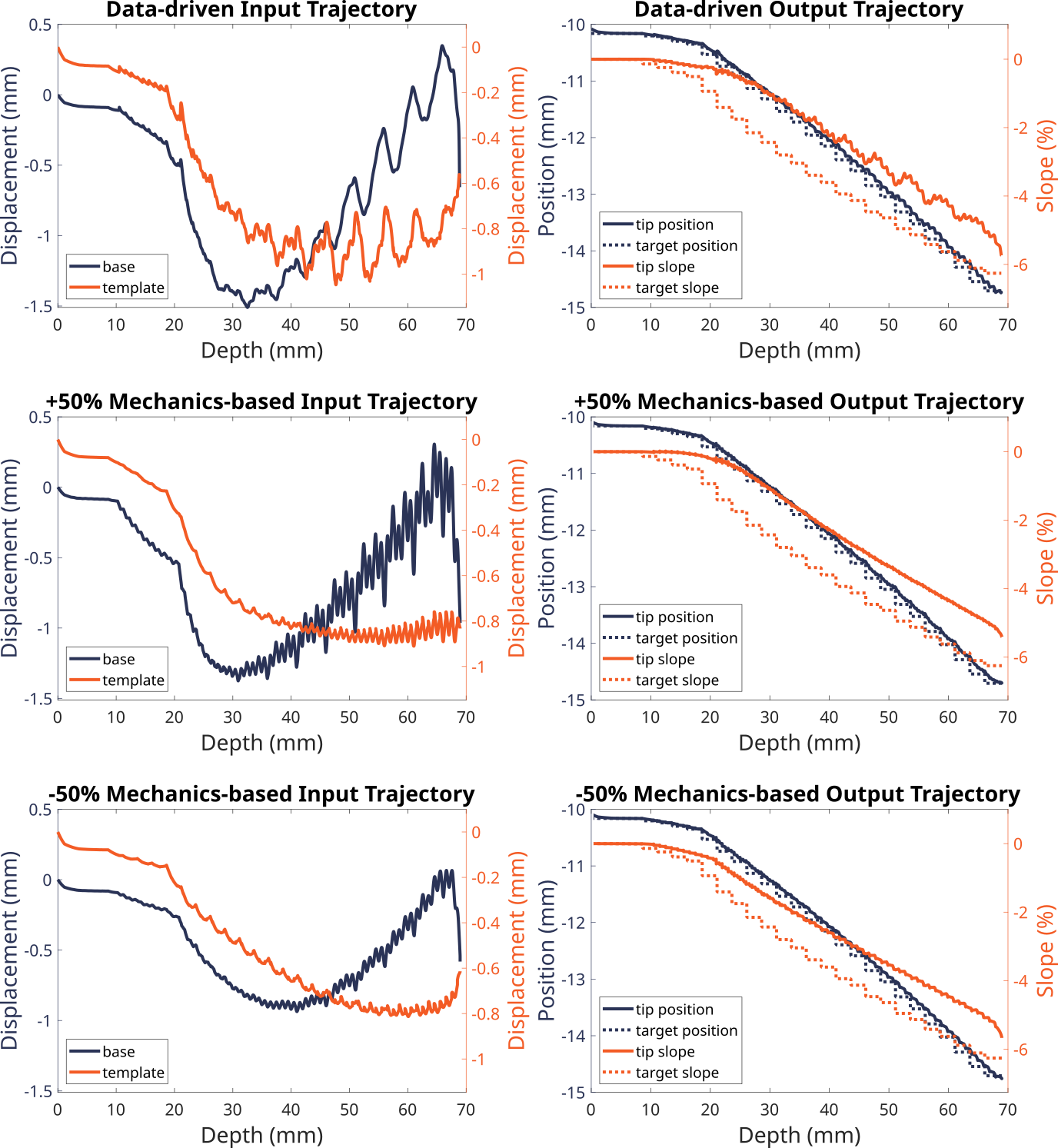}
  \caption{Needle tip and control input trajectories of path following task for Target 9. The $\pm 50\%$ represents parametric variations for the mechanics-based controller.}
  \label{fig:T9_path}
\end{figure}

\subsubsection{Path Following}
\label{sec:path_following}

In simulated experiments, both controllers are able to drive the needle tip to follow desired paths for all 12 targets without significant performance difference. We pick Target 9 as an illustrative example since it requires the largest insertion distance. As shown in Fig.~\ref{fig:T9_path}, with $\pm 50\%$ parametric variation, the mechanics-based controller is able to keep up with its data-driven counterpart, as both controllers keep the needle tip relatively close to the given path from entry to target.

Comparing the mechanics-based controllers in Fig.~\ref{fig:T9_path}, when the parameter values are increased by $50\%$, the controller perceives the tissues to be stiffer than they actually are. Therefore to generate the desired tip motion, larger manipulation efforts are needed, and vice versa. The sawtooth appearance of input signals is in part due to the discretization of the continuous needle path.

\subsubsection{Point Stabilization}
\label{sec:point_stabilization}

\begin{figure}[t]
  \centering
  \includegraphics[width=\columnwidth]{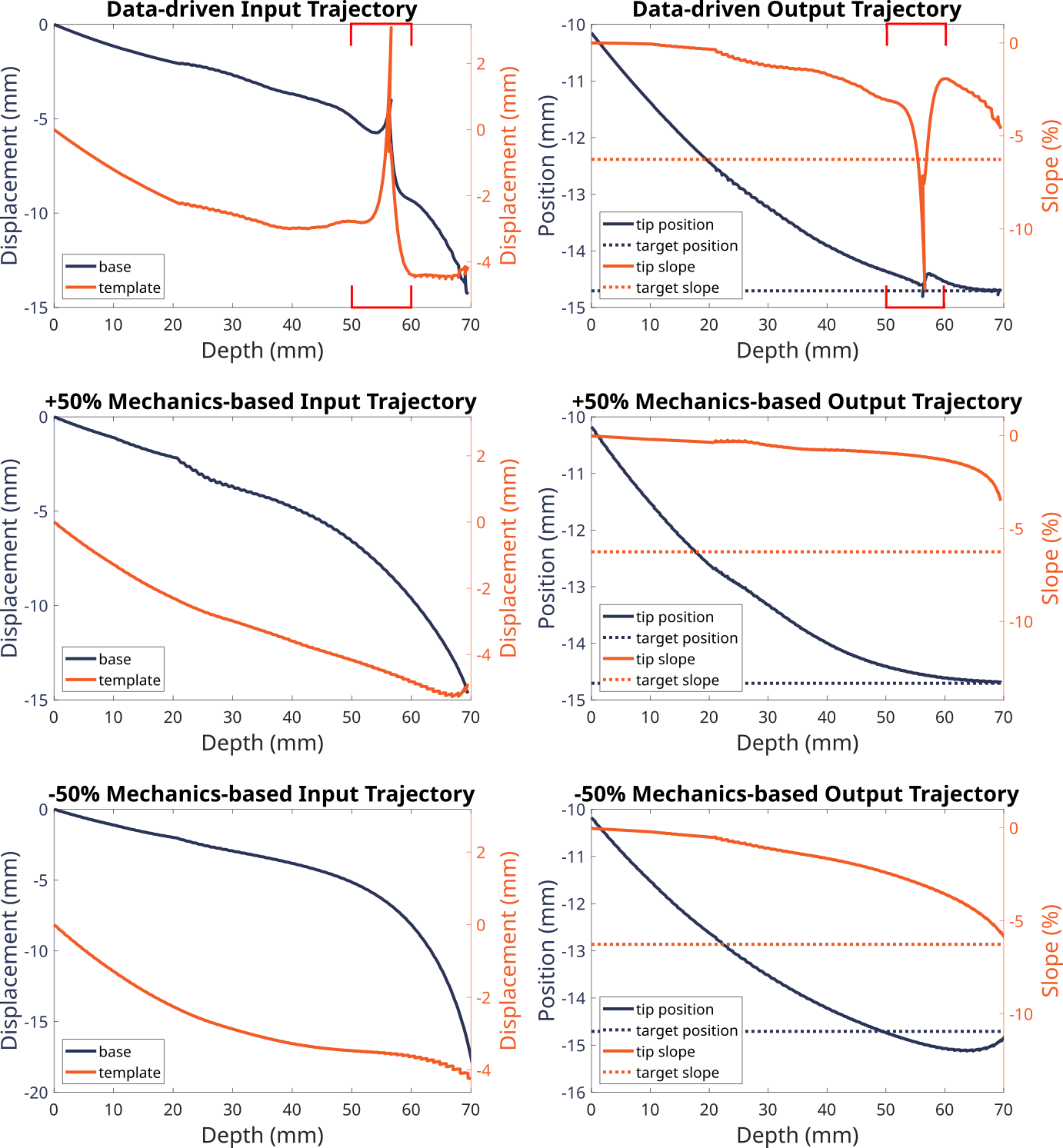}
  \caption{Needle tip and control input trajectories of point stabilization task for Target 9. The $\pm 50\%$ represents parametric variations for the mechanics-based controller. Red brackets emphasize the needle's ``pivoting'' behavior due to excessive tissue compression.}
  \label{fig:T9_pt}
\end{figure}

In simulated experiments, although both controllers can still reach all 12 target points, the difference between data-driven and mechanics-based controllers becomes evident when the path is removed and only entry and final target configurations are given.

Using Target 9 again as an example, shown in the top two graphs of Fig.~\ref{fig:T9_pt}, when the needle is inserted to around $55$mm, an increasingly large template manipulation is returned by the data-driven controller as the needle tip starts to overshoot from the desired needle deflection. Since the bevel direction is in the $-y$-direction, the controller attempts to correct the error by moving the needle shaft in the $+y$-direction such that by compressing the tissue, the controller is able to bring the needle tip upward. This strategy works well initially because tissue compression is small, and so is the resistance force; yet as the tissue is further compressed, strain-hardening tissue response increases the reaction force on the needle shaft, and such strategy quickly becomes ineffective.

Since bending of the needle still reduces tip error, albeit in a diminishing manner, inversion of the Jacobian mapping generates increasingly large control signals, shown as the rising input signal on the top left figure. Such template motion generates a ``pivoting'' behavior as the needle tip starts to point further down in the bevel direction, which can only result in further deviation from desired target as the needle inserts. This period is delimited by red brackets in Fig.~\ref{fig:T9_pt}.

To prevent the data-driven controller from generating excessive tissue compression, we implement a thresholding strategy that scales down the signal to regulate control inputs, and as a way to let the controller better observe the needle behavior under small adjustment before computing the next input signals. This strategy successfully limits the magnitude of the control signa, and allows the controller to complete the task. Other techniques can also be implemented to regulate the data-driven control signals such that the commanded shape manipulation falls within a safety margin and hardware limit.

In comparison, the mechanics-based controller generates smoother signals due to the fact that the system Jacobian is simulated by small variations of input signals, as explained previously with~\Cref{eq:central_difference}.

\begin{table*}[]
\caption{Summary of controller performance under two tasks, sorted in ascending order of target depth.}
\label{tab:summary}
\resizebox{\textwidth}{!}{%
\begin{tabular}{cccc|cccccc|cccccc|}
\cline{5-16}
 &  &  &  & \multicolumn{6}{c|}{Path-Following $P$ {[}\%{]}} & \multicolumn{6}{c|}{Point-Stabilization $P$ {[}\%{]}} \\ \hline
\multicolumn{1}{|c|}{\multirow{2}{*}{\begin{tabular}[c]{@{}c@{}}Target\\ Number\end{tabular}}} & \multicolumn{1}{c|}{\multirow{2}{*}{\begin{tabular}[c]{@{}c@{}}Depth\\ {[}mm{]}\end{tabular}}} & \multicolumn{1}{c|}{\multirow{2}{*}{\begin{tabular}[c]{@{}c@{}}Deflection\\ {[}mm{]}\end{tabular}}} & \multirow{2}{*}{\begin{tabular}[c]{@{}c@{}}Slope\\ {[}\%{]}\end{tabular}} & \multicolumn{2}{c|}{Data-driven} & \multicolumn{2}{c|}{+50\% Mech} & \multicolumn{2}{c|}{-50\% Mech} & \multicolumn{2}{c|}{Data-driven} & \multicolumn{2}{c|}{+50\% Mech} & \multicolumn{2}{c|}{-50\% Mech} \\ \cline{5-16} 
\multicolumn{1}{|c|}{} & \multicolumn{1}{c|}{} & \multicolumn{1}{c|}{} &  & \multicolumn{1}{c|}{base} & \multicolumn{1}{c|}{template} & \multicolumn{1}{c|}{base} & \multicolumn{1}{c|}{template} & \multicolumn{1}{c|}{base} & template & \multicolumn{1}{c|}{base} & \multicolumn{1}{c|}{template} & \multicolumn{1}{c|}{base} & \multicolumn{1}{c|}{template} & \multicolumn{1}{c|}{base} & template \\ \hline
\multicolumn{1}{|c|}{1} & \multicolumn{1}{c|}{30.62} & \multicolumn{1}{c|}{-1.00} & -2.00 & \multicolumn{1}{c|}{3.74} & \multicolumn{1}{c|}{1.98} & \multicolumn{1}{c|}{3.74} & \multicolumn{1}{c|}{1.98} & \multicolumn{1}{c|}{2.38} & 1.48 & \multicolumn{1}{c|}{6.00} & \multicolumn{1}{c|}{2.82} & \multicolumn{1}{c|}{8.41} & \multicolumn{1}{c|}{3.36} & \multicolumn{1}{c|}{4.43} & 2.76 \\ \hline
\multicolumn{1}{|c|}{2} & \multicolumn{1}{c|}{33.02} & \multicolumn{1}{c|}{-1.21} & -2.00 & \multicolumn{1}{c|}{4.12} & \multicolumn{1}{c|}{2.31} & \multicolumn{1}{c|}{4.36} & \multicolumn{1}{c|}{2.35} & \multicolumn{1}{c|}{2.54} & 1.61 & \multicolumn{1}{c|}{7.01} & \multicolumn{1}{c|}{3.05} & \multicolumn{1}{c|}{9.19} & \multicolumn{1}{c|}{3.72} & \multicolumn{1}{c|}{5.23} & 3.05 \\ \hline
\multicolumn{1}{|c|}{3} & \multicolumn{1}{c|}{35.42} & \multicolumn{1}{c|}{-1.38} & -3.00 & \multicolumn{1}{c|}{3.49} & \multicolumn{1}{c|}{1.76} & \multicolumn{1}{c|}{3.69} & \multicolumn{1}{c|}{1.76} & \multicolumn{1}{c|}{2.04} & 1.09 & \multicolumn{1}{c|}{6.90} & \multicolumn{1}{c|}{1.79} & \multicolumn{1}{c|}{9.04} & \multicolumn{1}{c|}{2.50} & \multicolumn{1}{c|}{5.28} & 1.83 \\ \hline
\multicolumn{1}{|c|}{4} & \multicolumn{1}{c|}{40.22} & \multicolumn{1}{c|}{-1.80} & -3.00 & \multicolumn{1}{c|}{3.07} & \multicolumn{1}{c|}{1.69} & \multicolumn{1}{c|}{3.01} & \multicolumn{1}{c|}{1.58} & \multicolumn{1}{c|}{1.88} & 1.09 & \multicolumn{1}{c|}{8.87} & \multicolumn{1}{c|}{2.21} & \multicolumn{1}{c|}{10.77} & \multicolumn{1}{c|}{3.10} & \multicolumn{1}{c|}{7.76} & 2.28 \\ \hline
\multicolumn{1}{|c|}{12} & \multicolumn{1}{c|}{40.22} & \multicolumn{1}{c|}{-1.80} & -3.00 & \multicolumn{1}{c|}{3.68} & \multicolumn{1}{c|}{2.27} & \multicolumn{1}{c|}{3.35} & \multicolumn{1}{c|}{1.94} & \multicolumn{1}{c|}{2.22} & 1.45 & \multicolumn{1}{c|}{9.42} & \multicolumn{1}{c|}{3.07} & \multicolumn{1}{c|}{11.32} & \multicolumn{1}{c|}{4.06} & \multicolumn{1}{c|}{8.39} & 3.17 \\ \hline
\multicolumn{1}{|c|}{11} & \multicolumn{1}{c|}{45.02} & \multicolumn{1}{c|}{-2.23} & -4.00 & \multicolumn{1}{c|}{3.12} & \multicolumn{1}{c|}{2.12} & \multicolumn{1}{c|}{2.94} & \multicolumn{1}{c|}{1.76} & \multicolumn{1}{c|}{1.99} & 1.41 & \multicolumn{1}{c|}{11.03} & \multicolumn{1}{c|}{3.04} & \multicolumn{1}{c|}{13.03} & \multicolumn{1}{c|}{4.43} & \multicolumn{1}{c|}{10.91} & 3.33 \\ \hline
\multicolumn{1}{|c|}{5} & \multicolumn{1}{c|}{54.62} & \multicolumn{1}{c|}{-3.08} & -5.00 & \multicolumn{1}{c|}{2.36} & \multicolumn{1}{c|}{1.56} & \multicolumn{1}{c|}{2.20} & \multicolumn{1}{c|}{1.33} & \multicolumn{1}{c|}{1.38} & 1.12 & \multicolumn{1}{c|}{15.03} & \multicolumn{1}{c|}{3.81} & \multicolumn{1}{c|}{16.28} & \multicolumn{1}{c|}{4.66} & \multicolumn{1}{c|}{16.95} & 3.25 \\ \hline
\multicolumn{1}{|c|}{6} & \multicolumn{1}{c|}{54.62} & \multicolumn{1}{c|}{-3.08} & -5.00 & \multicolumn{1}{c|}{2.35} & \multicolumn{1}{c|}{1.63} & \multicolumn{1}{c|}{2.25} & \multicolumn{1}{c|}{1.35} & \multicolumn{1}{c|}{1.42} & 1.17 & \multicolumn{1}{c|}{14.73} & \multicolumn{1}{c|}{3.28} & \multicolumn{1}{c|}{16.27} & \multicolumn{1}{c|}{4.80} & \multicolumn{1}{c|}{16.92} & 3.39 \\ \hline
\multicolumn{1}{|c|}{10} & \multicolumn{1}{c|}{61.82} & \multicolumn{1}{c|}{-3.78} & -6.00 & \multicolumn{1}{c|}{2.28} & \multicolumn{1}{c|}{1.70} & \multicolumn{1}{c|}{2.31} & \multicolumn{1}{c|}{1.52} & \multicolumn{1}{c|}{1.57} & 1.37 & \multicolumn{1}{c|}{17.96} & \multicolumn{1}{c|}{5.17} & \multicolumn{1}{c|}{18.74} & \multicolumn{1}{c|}{6.49} & \multicolumn{1}{c|}{21.47} & 4.45 \\ \hline
\multicolumn{1}{|c|}{7} & \multicolumn{1}{c|}{64.22} & \multicolumn{1}{c|}{-4.07} & -6.00 & \multicolumn{1}{c|}{1.96} & \multicolumn{1}{c|}{1.33} & \multicolumn{1}{c|}{1.88} & \multicolumn{1}{c|}{1.14} & \multicolumn{1}{c|}{1.18} & 0.99 & \multicolumn{1}{c|}{25.04} & \multicolumn{1}{c|}{\textbf{29.69}} & \multicolumn{1}{c|}{19.87} & \multicolumn{1}{c|}{6.38} & \multicolumn{1}{c|}{23.51} & 4.08 \\ \hline
\multicolumn{1}{|c|}{8} & \multicolumn{1}{c|}{66.62} & \multicolumn{1}{c|}{-4.28} & -6.00 & \multicolumn{1}{c|}{1.92} & \multicolumn{1}{c|}{1.31} & \multicolumn{1}{c|}{1.84} & \multicolumn{1}{c|}{1.14} & \multicolumn{1}{c|}{1.16} & 0.98 & \multicolumn{1}{c|}{27.43} & \multicolumn{1}{c|}{\textbf{42.07}} & \multicolumn{1}{c|}{20.73} & \multicolumn{1}{c|}{6.78} & \multicolumn{1}{c|}{25.25} & 4.97 \\ \hline
\multicolumn{1}{|c|}{9} & \multicolumn{1}{c|}{69.02} & \multicolumn{1}{c|}{-4.55} & -6.00 & \multicolumn{1}{c|}{2.20} & \multicolumn{1}{c|}{1.52} & \multicolumn{1}{c|}{2.00} & \multicolumn{1}{c|}{1.32} & \multicolumn{1}{c|}{1.36} & 1.18 & \multicolumn{1}{c|}{20.76} & \multicolumn{1}{c|}{6.56} & \multicolumn{1}{c|}{21.29} & \multicolumn{1}{c|}{7.67} & \multicolumn{1}{c|}{26.12} & 6.15 \\ \hline
\end{tabular}%
}
\end{table*}

\subsection{Task Performance Comparison}
\label{sec:task_performance_comparison}
It is insightful to use the performance metric $P$ from~\Cref{eq:metric} to further compare the two tasks. For path-following, the magnitude of shape manipulation is overall smaller compared to point-stabilization. More specifically, the performance metric values for path-following are below $5\%$ for all targets regardless of insertion depth; however, for point-stabilization, the $P$ values appear to increase with deeper targets, suggesting a larger magnitude of needle shape change needed for deeper insertions.

\begin{figure}[t]
  \centering  \includegraphics[width=\columnwidth]{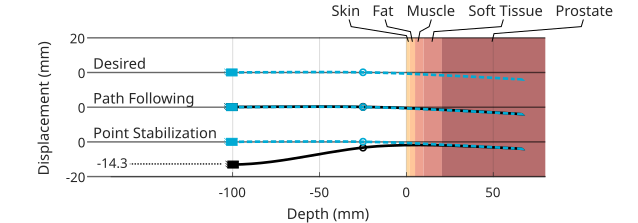}
  \caption{Target 9 needle shapes comparison against the desired needle shape. Data-driven controller is used for path following (Fig.~\ref{fig:T9_path} top) and point stabilization (Fig.~\ref{fig:T9_pt}).}
  \label{fig:final_shape_comparison}
\end{figure}

This behavior can also be observed directly by comparing Fig.~\ref{fig:T9_path} and Fig.~\ref{fig:T9_pt}: to reach the same target, the shape change required to follow the path is on the magnitude of around 1mm, but can require ten times the effort to drive the needle when no path is provided. Under point-stabilization, two instances can be found in Table~\ref{tab:summary} where the data-driven controller generates a maximum needle template displacement of over $20\%$ of the insertion depth, suggesting that over $20$mm displacement is required near the entry point, which can lead to significant risk of tissue damage.

This is made clear in Fig.~\ref{fig:final_shape_comparison} when comparing the final needle shapes for the two tasks to the desired needle shape, from which the needle path is generated. The needle shape after point stabilization (top graphs of Fig.~\ref{fig:T9_pt}) deviates from the desired shape, particularly at the needle base where a large vertical displacement is generated and results in significant needle shape change; for path following (top graphs of Fig.~\ref{fig:T9_path}), the resulting shape nearly overlaps with the desired shape.

The difference in $P$ values for path-following and point-stabilization tasks highlights the importance of using a feasible needle path, particularly for numerical-Jacobian-based resolved-rate controllers for needle control. Since numerical Jacobians only capture the system behavior in a small neighborhood, presenting the controllers with intermediate goal points can loosen the requirement of accurate system dynamics over a long time horizon. In the case of autonomous bevel-tip needle insertion under shape manipulation, this can mean the difference between a successful insertion and a irreversible damage to the soft tissue.

The two highlighted $P$ values for the data-driven controller can be mitigated by placing a heavier restriction on the input signals such that needle ``pivoting'' is avoided completely; however, our current goal is to demonstrate the controller behavior and suggest a possible solution.

Note that in \Cref{eq:metric}, $P$ is scaled by the depth of each target $x_{target}$. This choice is motivated by our intuition that smaller needle manipulations can lead to a safer operation. Other metrics can be used that take into account the rate of needle manipulation, as well as the depth at which maximum manipulation is applied. A more thorough investigation of rate-dependent soft tissue behaviors under large strain, such as tissue tearing, is needed to better motivate a performance metric that correlates more directly with patient safety. 



\section{Conclusions and Future Work}
\label{sec:conclusion_and_future_work}
In this paper, we compare two resolved-rate controllers in a simulator for bevel-tip needle control with shape manipulation. Although both data-driven and mechanics-based controllers can successfully reach $12/12$ target points, the mechanics-based controller can generate physically-informed control signals under parametric uncertainty, while signal regulation is needed for the data-driven controller to ensure successful completion of the control task. Both controllers work better when provided a path leading to the target, but the mechanics-based controller is better behaved when only final target positions are provided.

Future work includes further investigation of the two controllers when the number of inputs mismatch the number of outputs, as well as integrating needle bending and insertion dynamics as part of the control algorithm. Safety margins for tissue compression will be investigated and incorporated in the control algorithm as well. Physical experiments will be carried out to further compare the two controllers in the presence of sensor noise.

\bibliography{refs}
\bibliographystyle{ieeetr}

\end{document}